\documentclass{article}
\usepackage{ijcai23}

\usepackage{times}
\usepackage{soul}
\usepackage{url}
\usepackage[hidelinks]{hyperref}
\usepackage[utf8]{inputenc}
\usepackage[small]{caption}
\usepackage{graphicx}
\usepackage{amsmath}
\usepackage{amsthm}
\usepackage{booktabs}
\usepackage{algorithm}
\usepackage{algorithmic}
\usepackage[switch]{lineno}
\usepackage{amssymb}
\usepackage{multirow}

\title{
Organizing Background to Explore Latent Classes for Incremental Few-shot Semantic Segmentation
}

\author{
Lianlei Shan$^1$
\and
Wenzhang Zhou$^2$\and
Wei Li$^{2,3}$\And
Xingyu Ding$^4$
\affiliations
$^1$University of Chinese Academy of Sciences\\
$^2$Nanjing University of Posts and Telecommunications\\
$^3$Beijing University of Posts and Telecommunications\\
$^4$University of Chinese Academy of Sciences
\emails
shanlianlei18@mails.ucas.edu.cn,
0230196@njupt.edu.cn,
leesoon@bupt.edu.cn,
dingxingyu21@mails.ucas.ac.cn
}

\begin{document}

\maketitle

\begin{abstract}
The goal of incremental Few-shot Semantic Segmentation (iFSS) is to extend pre-trained segmentation models to new classes via few annotated images without access to old training data. During incrementally learning novel classes, the data distribution of old classes will be destroyed, leading to catastrophic forgetting. Meanwhile, the novel classes have only few samples, making models impossible to learn the satisfying representations of novel classes. For the iFSS problem, we propose a network called OINet, i.e., the background embedding space \textbf{O}rganization and prototype \textbf{I}nherit Network. Specifically, when training base classes, OINet uses multiple classification heads for the background and sets multiple sub-class prototypes to reserve embedding space for the latent novel classes. During incrementally learning novel classes, we propose a strategy to select the sub-class prototypes that best match the current learning novel classes and make the novel classes inherit the selected prototypes' embedding space. This operation allows the novel classes to be registered in the embedding space using few samples without affecting the distribution of the base classes. Results on Pascal-VOC and COCO show that OINet achieves a new state of the art.
\end{abstract}

\section{Introduction}
\begin{figure}[H]
\centering
\includegraphics[scale=0.185]{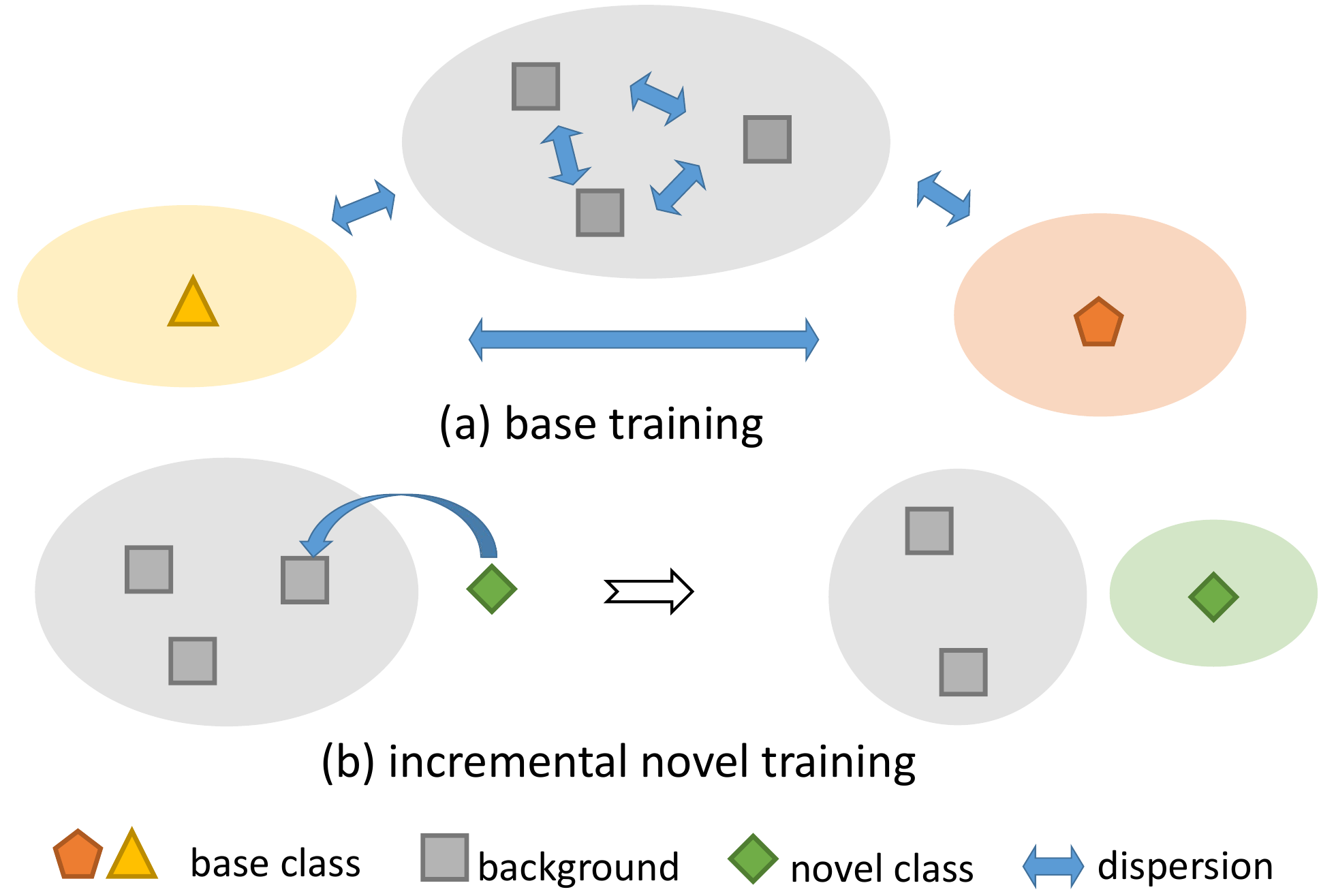}
\caption{
Illustration of the proposed method.
(a) denotes the learning of base classes. Multiple sub-class prototypes are set for the background. Through embedding space organization, the sub-class prototypes and base classes are dispersed from each other.
(b) represents the incremental learning of novel classes.
First, the prototypes that best match the novel classes are selected from the sub-class prototypes of the background, and then novel classes are made to inherit the embedding spaces of the selected sub-class prototypes.
Therefore, the whole process not only ensures that the novel classes can be learned (i.e., registered to the embedding space) but also makes the learning of novel classes not affect the distribution of base classes.
}
\label{ifss_intro}
\end{figure}
\vspace{-0.1cm}

Incremental semantic segmentation \cite{mib,plop} can continuously learn new classes without losing the ability to segment old classes, thus improving the adaptability of models.
In the real world, due to the cost of expensive pixel-level annotations for semantic segmentation tasks, we expect the model to learn new classes with only few samples. 
The task of incrementally learning new classes through small samples is called incremental Few-shot Semantic Segmentation (iFSS).
Compared with incremental segmentation, the samples of novel classes in iFSS are very limited (generally less than 5 of each class), so segmentation models are difficult to acquire correct and complete representations of novel classes.
Moreover, same with other incremental learning tasks, catastrophic forgetting of the old classes is also a challenging problem.
Few-shot Semantic Segmentation (FSS) \cite{fss} learns to segment new classes from few images but completely discards old knowledge. In contrast, Generalized FSS (GFSS) methods \cite{gfss} segment old and new classes but require access to training data for old classes, which is impossible in some cases, for example, if the model is on a device with limited storage.
Compared with them, iFSS aims to learn a segmentation model for both base and novel classes with only few new samples and does not require access to old data.

Based on the above analysis, the main challenges of iFSS are the catastrophic forgetting of the acquired knowledge and the difficulty of obtaining complete and accurate representations of novel classes with few samples.
Therefore, an effective iFSS framework should satisfy: 1) when learning new classes, it does not affect the feature distribution of old classes to avoid catastrophic forgetting, and 2) the representations of new classes can be learned with few samples, i.e., the features of the new classes can be registered in the embedding space with only few samples.
In order to meet the above two requirements, in this paper, we propose OINet, i.e., the background embedding space \textbf{O}rganization and prototype \textbf{I}nherit Network for iFSS.
iFSS can be divided into the initial learning of base classes and the subsequent multi-step incremental few-shot learning of novel classes.
The main idea of OINet is to reserve space for novel classes when learning base classes;
more importantly, the novel classes can inherit part of the space of background, so it is unnecessary to re-open (learning) new spaces, thus reducing the requirement of training samples.

Figure \ref{ifss_intro} succinctly shows the idea of our work. The main idea is the embedding space organization during base training and inheritance during learning novel classes.
Specifically, since the base classes have sufficient training samples, base training is critical for the construction of the embedding space.
Previous methods \cite{pifs,ifss_2} classify other classes (except base classes) as background, causing the embedding space to be basically allocated by each base class and background. Meanwhile, the features of various classes in the background are mixed in the embedding space.
These operations make novel classes easily occupy the space of the base classes during incremental learning. Meanwhile, it is also difficult to separate the novel class features from mixed background spaces with few samples.
Therefore, when learning base classes, the space for novel classes should be reserved from the background. We set multiple classification heads for the background (which can be considered as multiple sub-class prototypes) when learning base classes, and use a variety of operations to disperse the sub-class prototypes to reserve space for novel classes, as shown in Figure \ref{ifss_intro} (a).
In this way, base classes and novel classes do not interfere with each other in the embedding space, thereby alleviating catastrophic forgetting.

When learning novel classes, due to the limited samples, it is difficult to train the network to open up new embedding spaces for the novel classes.
Therefore, we select the sub-class prototypes that best match novel classes through off-the-shelf matching algorithms.
And we use the selected prototypes to obtain the classification heads of novel classes so that the novel classes can inherit the embedding spaces of the sub-class prototypes, as shown in Figure \ref{ifss_intro} (b).
This method makes novel classes not occupy or invade the embedding space of base classes but can make novel classes registered in the embedding space with limited samples.

In summary, our contributions are as follows,
\begin{itemize}
\item According to the characteristic of iFSS, we design a framework that bridges the base and novel training stages, which meets the requirements of iFSS and achieves significant improvements on benchmark datasets.
\item During base training, the background is set by multiple classification heads, from which sub-class prototypes are established. A variety of methods are proposed to disperse sub-class prototypes to reserve space for novel classes.
\item 
Since the novel classes contain limited samples, we set up a strategy to make the novel classes inherit the embedding space of sub-class prototypes produced from base training. Thus, novel classes can be registered into the embedding space at a small cost, and will not violate the embedding space of base classes.
\end{itemize}

\section{Related Work}
\subsection{Incremental Few-shot Semantic Segmentation}
PIFS \cite{pifs} is the first work of iFSS. PIFS is based on prototypes and combines distillation and re-norm methods, which achieves outstanding performance compared with other FSS or incremental learning (IL) methods. Our work is based on PIFS.
\cite{ifss_2} maintains the old knowledge through super-class, but it needs to store old knowledge and support images, which keeps totally different settings from PIFS and ours.
The prototype-based methods are the mainstream. 

\subsection{Incremental Learning}
Incremental learning aims to extend the knowledge of the model without forgetting \cite{il_1}. This problem has been extensively studied in image classification \cite{il_cls_1,lwf,icarl,semantic_dfit} and, more recently, segmentation \cite{mib,plop,ilt,sdr}.
For incremental semantic segmentation, \cite{mib} noticed background shift and proposed modified classification and distillation losses.
\cite{sdr} adopted the most basic intra-class aggregation and inter-class scatter operation, but it did not consider the background change or mine the background information.

\subsection{Few-shot Learning}
 Few-shot Learning methods can be divided into optimization-based \cite{op_1,op_222,op_333} and metric learning-based \cite{metric_1,metric_2,wi}. \cite{dwi,metric_44} learned to extract per-class prototypes from few images via meta-learning. \cite{wi} proposed weight imprinting to add new class weights to cosine classifiers. \cite{metric_1} fixed the feature extractor and trained a classifier for the new class. \cite{fss_8} extended the segmentation task of \cite{metric_44} by aggregating pixel-wise feature representations. \cite{amp} proposed updating old classes when computing new class prototypes.
 PIFS \cite{pifs} learned an embedding space where instances of the same classes are close to each other. Unlike them, our method explores the information in the background when generating classifiers for new classes.

\begin{figure*}[htbp]
\centering
\includegraphics[scale=0.255]{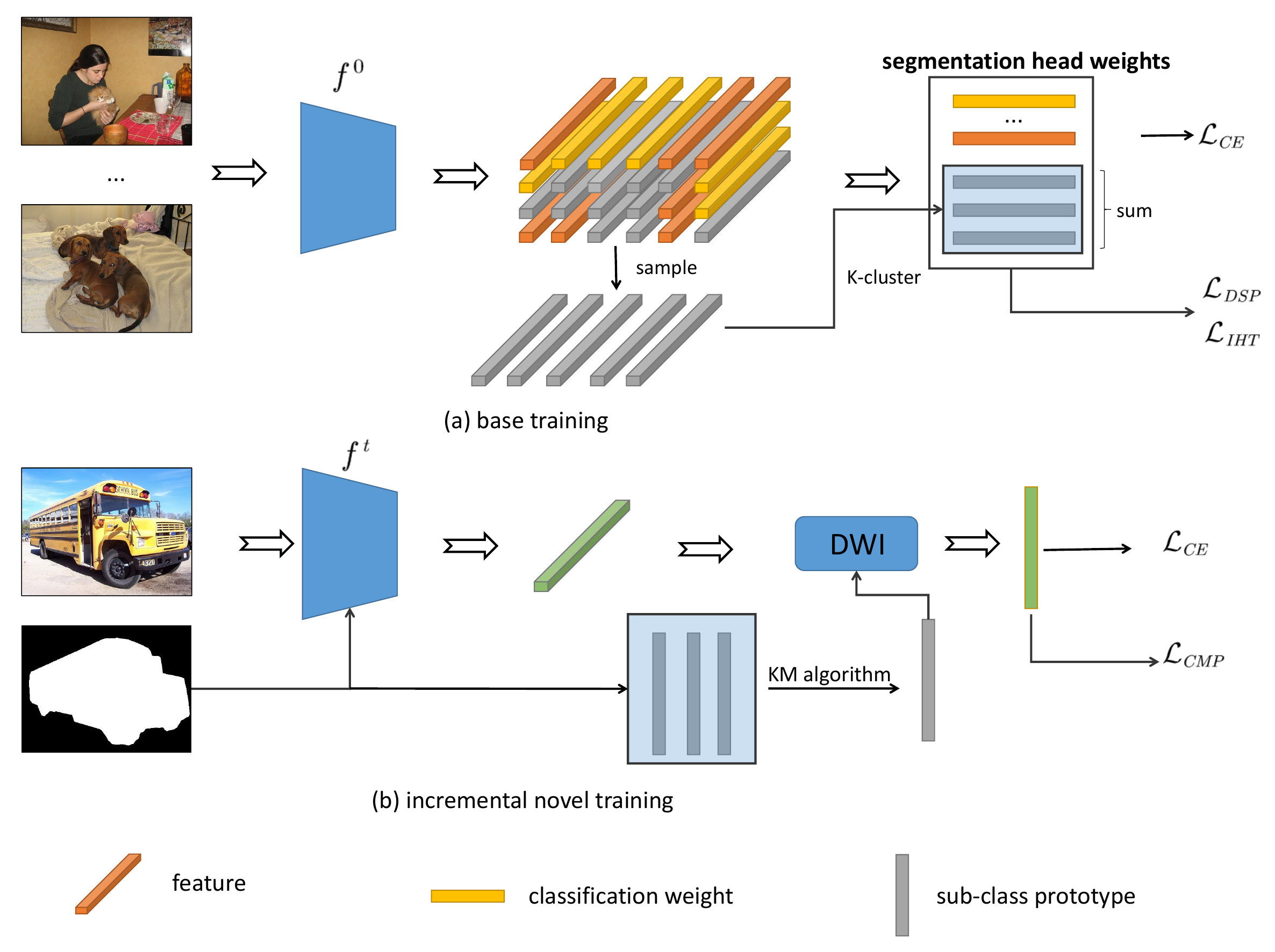}
\caption{
The overall structure of the proposed method.
The overall process can be divided into the stage of learning base classes, as shown in (a), and the stage of incrementally learning novel classes, as shown in (b).
The colors in the figure are consistent with those in Figure \ref{ifss_intro}, i.e., orange and yellow represent the base classes, gray represents the background, and green represents the currently learning novel class.
In base training, the difference is there are multiple classification heads for background (like $K$), and the weights of these classification heads are obtained from the features of the background through the k-cluster.
In incremental novel training, the weights of the segmentation heads of the novel classes are calculated through the training data of the novel classes and the sub-class prototypes obtained during base training.
DWI denotes dynamic weight imprinting.
}
\label{pic_overview}
\end{figure*}
\vspace{-0.1cm}

\section{Task Definition}
Formally, we denote the set of semantic classes already learned after step $t$ as $\mathcal{C}^{0:t}$, where a learning step denotes an update of the model's output space. During training, the network receives a sequence of data $\left\{\mathcal{D}^0, \ldots, \mathcal{D}^T\right\}$, where $\mathcal{D}^t=\{(x, y) \mid x \in$ $\left.\mathcal{X}, y \in \mathcal{Y}^t\right\}$. 
In each $\mathcal{D}^t$, $x$ is an image in the space $\mathcal{X} \in \mathbb{R}^{|I| \times 3}$, with $I$ the set of pixels, and $y$ its corresponding label mask in $\mathcal{Y}^t \subset\left(C^t\right)^{|I|}$.
Note that $\mathcal{D}^0$ is a large dataset while $\mathcal{D}^t$ are few-shot ones, i.e. $\left|\mathcal{D}^0\right| \gg\left|\mathcal{D}^t\right|, \forall t \geq 1$.
The model is first trained on the large dataset $\mathcal{D}^0$ and incrementally updated with few-shot datasets. We name the first learning step on $\mathcal{D}^0$ as the base step. Note that at step $t$, the model has access only to $\mathcal{D}^t$.
For $\mathcal{D}^t$, if an image contains the class $\mathcal{C}^{t}$, the image is selected. The difference is that except $\mathcal{C}^{t}$, the other classes on the label are masked as the background. 

Our goal is to learn a model $\phi^t$ that maps each pixel to a probability distribution over the set of classes, i.e., $\phi^t: \mathcal{X} \rightarrow \mathbb{R}^{|I| \times\left|\mathcal{C}^{0:t}\right|}$, where $t$ denotes the last learning step. We assume $\phi^t$ composed of a feature extractor $f^t: \mathcal{X} \rightarrow \mathbb{R}^{|I| \times d}$ and a classifier $g^t: \mathbb{R}^{|I| \times d} \rightarrow \mathbb{R}^{|I| \times\left|\mathcal{C}^{0:t}\right|}$, such that $\phi^t=g^t \circ f^t$. Here, $d$ is the feature dimension and $g^t$ is a classifier with parameters $\boldsymbol{W}^t=\left[\boldsymbol{w}_{0}^t, \ldots, \boldsymbol{w}_{\left|\mathcal{C}^t\right|}^t\right] \in \mathbb{R}^{d \times\left|\mathcal{C}^{0:t}\right|}$.
\vspace{-0.1cm}
\section{Method}
\subsection{Overview}
The overall framework of our method is shown in Figure \ref{pic_overview}, which can be divided into the stage of learning base classes, as in Figure \ref{pic_overview} (a), and the stage of incrementally learning novel classes, as in Figure \ref{pic_overview} (b).
During base training, consistent with the common segmentation network training, the image enters the feature extraction module $f^0$ and the classifier $g^0$ to calculate $\mathcal{L}_{ce}$. The difference is that we set multiple classification heads for the background class, and the weights of the multiple classification heads are determined through the features of the background class after the cluster. Each classification head of the background can be seen as a sub-class prototype. We organize the embedding space according to sub-class prototypes and segmentation weights of base classes to disperse each class and reserve space for novel classes. The details are introduced in Section \ref{sec_embedding_space_org}.

In the incrementally novel training stage, the task is to register the novel classes into the embedding space without affecting the distribution of the base classes. We select one or more sub-class prototypes that best match the feature of a novel class. Then the novel class is allowed to inherit the embedding space of the selected sub-class prototypes, i.e., the features in the embedding space are classified as the novel class.
As shown in Figure \ref{pic_overview} (b), the output obtained by incremental novel training is the segmentation head weights of the current learning novel classes.
We use the Kuhn Munkres (KM) algorithm to select the most appropriate sub-class prototype, and Intersection-over-Union (IoU) with true label masks is used to calculate the cost matrix.
We use one modified head generator (DWI in Figure \ref{pic_overview}) to generate the weights of novel classes.
And the Mean Squared Error (MSE) loss is used to make the features of the novel classes closer to the selected prototypes.
Specific details are presented in Section \ref{sec_select_inher}.

\subsection{Embedding Space Organization}
\label{sec_embedding_space_org}
In order to obtain the sub-class prototypes of the background that are not clustering or contain overlaps, we adopt multiple optimization objectives to organize embedding space. Among them, pixel-prototype contrastive learning is employed to disperse sub-class prototypes and base classes (i.e., inter-class dispersion) and pixel-prototype distance optimization to aggregate the features within the clusters (i.e., intra-class compaction).

\textbf{Inter-class Dispersion.}
We cluster the features of the background
into $K$ prototypes $\left\{\boldsymbol{p}_{ k}\right\}_{k=1}^K$. After all the samples in the current batch are processed, each pixel (feature) $\boldsymbol{i}$ belonging to the background is assigned to $k_i$-th prototype; thus, we can get the sub-class label of each background pixel.
Besides, we can also get the base class features and the segmentation head weights (which can be regarded as the prototypes of base classes).
Thus, it is natural to derive an optimization objective for prototype assignment prediction, i.e., maximize the prototype assignment posterior probability. 
This can be viewed as a pixel-prototype contrastive learning-based loss $\mathcal{L}_{\mathrm{DSP}}$, which is shown as,
\begin{equation}
\mathcal{L}_{\mathrm{DSP}}=-\log \frac{\exp \left(\boldsymbol{i}^{\top} \boldsymbol{p}_{i} / \tau\right)}{\exp \left(\boldsymbol{i}^{\top} \boldsymbol{p}_{i} / \tau\right)+\sum_{p^{-} \in \mathcal{P}^{-}} \exp \left(\boldsymbol{i}^{\top} \boldsymbol{p}^{-} / \tau\right)},
\end{equation}
where $\boldsymbol{p}_{i}$ denotes the prototype of embedding $\boldsymbol{i}$, and 
$\mathcal{P}^{-}=\left\{\boldsymbol{p}_{c}\right\}_{c=1}^{\left|\mathcal{C}^{0:t}\right|} \cup \left\{\boldsymbol{p}_{k}\right\}_{k=1}^{K} / \boldsymbol{p}_{i}$, which denotes the set of negative samples.
Intuitively, $\mathcal{L}_{\mathrm{DSP}}$ enforces each pixel embedding $\boldsymbol{i}$ to be similar with its assigned positive prototype $\boldsymbol{p}_{i}$, and dispersed with other $\left|\mathcal{C}^{0:t}\right|+K-1$ irrelevant and negative prototypes $\mathcal{P}^{-}$. 
In this way, not only the embeddings of the base classes can be dispersed, but more importantly, the sub-class prototypes of the background can also be distinguished from each other. The complete dispersion allows each embedding space of the background to be explored, which facilitates the registration (learning) of new classes.
The temperature $\tau$ (default is 0.1) controls the concentration level of representations.  
In contrast to prototype-based between-class and within-class scatter matrices, this method does not conduct heavy pixel pair-wise comparison, so the computational complexity is reduced.

\textbf{Intra-class Compaction.}
Based on the comparison over pixel-prototype distances, $\mathcal{L}_{\mathrm{DSP}}$ inspires inter-class (cluster) discrimination, but less considers reducing the intra-cluster variation, i.e., making pixel features of the same prototype compact. Therefore, a compactness-driven loss is needed for further regularizing representations by directly minimizing the distance between each embedded pixel and its assigned prototype. Inspired by \cite{protoseg}, the intra-class compaction loss $\mathcal{L}_{\text {CMP}}$ is as follows,
\begin{equation}
\mathcal{L}_{\text {CMP}}=\left(1-\boldsymbol{i}^{\top} \boldsymbol{p}_{i}\right)^2,
\end{equation}
where $\boldsymbol{p}_{i}$ is the sub-class prototype or the normalized segmentation weight corresponding to feature $\boldsymbol{i}$.
It should be noted that both $\boldsymbol{i}$ and $\boldsymbol{p}_{i}$ are $\ell_2$-normalized. $\mathcal{L}_{\text {CMP}}$ minimizes intra-cluster variations while $\mathcal{L}_{\mathrm{DSP}}$ maintains dispersion between features with different prototype assignments, making the embedding space more suitable for the representation learning of novel classes.

\textbf{Sub-class Prototype Update.}
The sub-class prototypes are obtained by clustering the features of the background. Meanwhile, these prototypes also serve as classification heads, i.e., these prototypes are also changed by stochastic gradient descent.
Specifically, we first cluster the background features to obtain sub-class prototypes and then use the prototypes to initialize multiple classification head weights of the background. 
When a new batch of images comes for training, the initial points of clusters are the classification head weights obtained in the previous iteration. In this way, the new cluster centers and the original sub-class prototypes can be in one-to-one correspondence.
Therefore, after each training iteration, each prototype is updated as follows:
\begin{equation}
\boldsymbol{p}_{k} \leftarrow \mu \boldsymbol{p}_{k}+(1-\mu) \overline{\boldsymbol{i}}_{k},
\label{eq_mu}
\end{equation}
where $\mu \in[0,1]$ is a momentum coefficient, and $\overline{\boldsymbol{i}}_{k}$ denotes the $\ell_2$-normalized mean vector of the embedded pixels assigned to prototype $\boldsymbol{p}_{k}$.
Prototypes are changed by gradient descent so that background can be separated from other base classes. Besides, they are also updated with the cluster centers so that various sub-class prototypes can also be distinguished from each other.

\subsection{Selection and Inheritance Mechanisms}
\label{sec_select_inher}
When incrementally learning novel classes, in order to not affect the distribution of base classes and enable the segmentation network to quickly learn the representations of novel classes with few samples, we design a selection strategy to choose the best-matched sub-class prototypes for novel classes and an inheritance (generation) strategy to generate weights of segmentation heads of novel classes.

\textbf{Sub-class prototype selection strategy.}
Given the sub-class prototypes and novel classes, we need to give them the best match, which can be obtained by the Kuhn-Munkres (KM) algorithm. The key to the KM algorithm is to construct the cost matrix. We use $X_{ij}$ to denote the cost value of the $i$-th novel class and the $j$-th prototype.
Specifically, given the training samples of a novel class $i$, we can obtain its true label mask $M_i$.
Meanwhile, using sub-class prototype $\boldsymbol{p}_{j}$ as the classifier and the features extracted from the novel class images as input, the model can output a binary mask $\hat{M}_{i_j}$ after normalization and quantization.
It should be noted that after determining the novel class $i$ and the input image, the binary masks $\hat{M}_{i_j}$ can be calculated for prototype $\boldsymbol{p}_{j}$.
The Intersection over Union (IoU) ${{IoU_{ij}}}$ can be computed for each $\hat{M}_{i_j}$ and $M_i$.
If ${{IoU_{ij}}}$ is large, the cost value $X_{ij}$ should be small.
Therefore, the cost value $X_{ij}$ can be defined as 1-${ {IoU_{ij}}}$.

Once the cost matrix is determined, the KM algorithm can determine the optimal matching, and every novel class will be assigned the most suitable prototypes.
The KM algorithm can be executed N times so that each novel class can obtain N prototypes.
When a sub-class prototype is selected, its parameters are initialized through a new clustering, and the parameters are still updated according to the original strategy.

\textbf{Novel class weights generation strategy.}
After obtaining the most suitable sub-class prototypes, we need to generate the classification heads of the novel classes.
The most direct way is to change the original sub-class prototypes to the segmentation head weights of the novel classes. In this way, the novel classes directly inherit the embedding space of the sub-class prototypes without change.
But in some classes, even the highest IoU is not satisfactory. Besides, we do not make full use of the precious training data of novel classes in this way.
So, a better strategy is to combine the selected prototypes and the features of novel classes. Inspired by \cite{dwi}, we introduce the dynamic weight imprint (DWI) strategy.
Since the operation is consistent at each incremental step, we omit the subscript $t$.
Thus, the novel class classification head $\boldsymbol{w}_{n}$ is calculated by
\begin{equation}
\boldsymbol{w}_{n}=\boldsymbol{w}_{f} \odot \boldsymbol{\overline{f}}_{n}+ \boldsymbol{w}_{att} \odot \boldsymbol{z}+ \boldsymbol{w}_p \odot \overline{\boldsymbol{p}}_{slc},
\end{equation}
where $\boldsymbol{\overline{f}}_{n}$ denotes the mean features of the current learning novel class, and $\overline{\boldsymbol{p}}_{slc}$ represents the selected sub-class prototype (take the mean value if prototypes are more than one).
$\boldsymbol{z}$ is the output of the base class classification heads after the same attention as in \cite{dwi}. Different from \cite{dwi}, the input newly adds the selected prototypes.
$\boldsymbol{w}_{f}$, $\boldsymbol{w}_{att}$, and $\boldsymbol{w}_p$ $\in \mathbb{R}^d$ are learnable weight vectors. $\odot$ is the Hadamard product.

In addition to the cross-entropy loss $\mathcal{L}_{\mathrm{CE}}$, we also add a restriction on the generated novel class weight to make $\boldsymbol{w}_{n}$ close to the embedding space of the selected sub-class prototypes, to achieve the purpose of inheritance, and the inheritance loss $\mathcal{L}_{IHT}$ is,
\begin{equation}
\mathcal{L}_{IHT}={\left\|\boldsymbol{w}_{n}-\overline{\boldsymbol{p}}_{slc}\right\|^2}.
\end{equation}

\subsection{Optimization Goal}
The whole training process is divided into base training and incremental training.
During base training, in addition to learning the good segmentation ability of the base class, the model also needs to organize the embedding space well. Therefore, the loss during base training is 
\begin{equation}
\mathcal{L}_{\mathrm{Base}}=\mathcal{L}_{\mathrm{CE}}+\lambda_1 \mathcal{L}_{\mathrm{DSP}}+\lambda_2 \mathcal{L}_{\text {CMP}}.
\end{equation}

During incremental learning, in addition to the loss of PIFS \cite{pifs} $\mathcal{L}_{\mathrm{PIFS}}$ (PIFS is our baseline), there are segmentation loss of novel classes $\mathcal{L}_{\mathrm{CE}}$ and inheritance loss  $\mathcal{L}_{\mathrm{IHT}}$, i.e.,
\begin{equation}
\mathcal{L}_{\mathrm{Novel}}=\mathcal{L}_{\mathrm{PIFS}}+\mathcal{L}_{\mathrm{CE}}+\lambda_3 \mathcal{L}_{\mathrm{IHT}}.
\end{equation}

\begin{table*}[htbp]
\centering
  \scalebox{0.75}{
   \begin{tabular}{cc|ccc|ccc|ccc}
   \toprule
   &\multirow{2}{*}{Method}&\multicolumn{3}{c|}{1-shot}&\multicolumn{3}{c|}{2-shot}&\multicolumn{3}{c}{5-shot}\\
   &&mIoU-Base& mIoU-Novel& HM&mIoU-Base& mIoU-Novel& HM&mIoU-Base& mIoU-Novel& HM\\
    \toprule
    &FT&47.2& 3.9& 7.2&53.5& 4.4& 8.1&58.7& 7.7& 13.6\\
    \midrule
    \multirow{3}{*}{\rotatebox{90}{FSC}}&WI&66.6& 16.1& 25.9& 66.6& 19.8& 30.5& 66.6& 21.9& 33.0\\
    &DWI&\textbf{67.2}& 16.3& 26.2& \textbf{67.5}& 21.6& 32.7& \textbf{67.6}& 25.4 &36.9\\
    &RT&49.2& 5.8 &10.4&36.0 &4.9& 8.6 &45.1 &10.0 &16.4\\
    \midrule
    \multirow{2}{*}{\rotatebox{90}{FSS}}&AMP&58.6& 14.5 &23.2& 58.4& 16.3& 25.5& 57.1& 17.2& 26.4\\
    &SPN&49.8& 8.1& 13.9& 56.4& 10.4& 17.6& 61.6& 16.3 &25.8\\
    \midrule
    \multirow{3}{*}{\rotatebox{90}{IL}}&LwF&42.1& 3.3 &6.2& 51.6& 3.9& 7.3 &59.8& 7.5 &13.4\\
    &ILT&43.7& 3.3 &6.1 &52.2& 4.4& 8.1& 59.0& 7.9 &13.9\\
    &MiB& 43.9& 2.6& 4.9& 51.9& 2.1& 4.0& 60.9& 5.8 &10.5\\
     \midrule
     &PIFS&64.1 &16.9& 26.7& 65.2& 23.7& 34.8 &64.5& 27.5& 38.6\\
    &OINet&66.1&\textbf{18.0}&\textbf{28.3}&66.3&\textbf{25.2}&\textbf{36.5}&66.4&\textbf{28.2}&\textbf{39.6}\\
  \bottomrule
\end{tabular}
}
\caption{Results on VOC.}
  \label{result_ifss_voc}
\end{table*}
\begin{table*}[htbp]
\centering
  \scalebox{0.75}{
   \begin{tabular}{cc|ccc|ccc|ccc}
   \toprule
   &\multirow{2}{*}{Method}&\multicolumn{3}{c|}{1-shot}&\multicolumn{3}{c|}{2-shot}&\multicolumn{3}{c}{5-shot}\\
   &&mIoU-Base& mIoU-Novel& HM&mIoU-Base& mIoU-Novel& HM&mIoU-Base& mIoU-Novel& HM\\
    \toprule
    &FT&38.5& 4.8 &8.6 &40.3& 6.8 &11.7& 39.5 &11.5& 17.8\\
    \midrule
    \multirow{3}{*}{\rotatebox{90}{FSC}}&WI&\textbf{46.3} &8.3 &14.0& \textbf{46.5} &9.3 &15.4& 46.3& 10.3& 16.8\\
    &DWI&46.2 &9.2 &15.3& \textbf{46.5}& 11.4 &18.3& \textbf{46.6}& 14.5& 22.1\\
    &RT&38.4 &5.2 &9.1& 43.8& 10.1& 16.4& 44.1& 16.0& 23.5\\
    \midrule
    \multirow{2}{*}{\rotatebox{90}{FSS}}&AMP&36.6 &7.9 &13.1 &36.0& 9.2 &14.6& 33.2& 11.0 &16.5\\
    &SPN&40.3 &8.7 &14.3 &41.7& 12.5& 19.2 &41.4 &18.2& 25.3\\
    \midrule
    \multirow{3}{*}{\rotatebox{90}{IL}}&LwF&41.0 &4.1 &7.4 &42.7 &6.5 &11.3 &42.3 &12.6 &19.4\\
    &ILT&43.7 &6.2 &10.8 &47.1& 10.0& 16.5& 45.3& 15.3& 22.8\\
    &MiB&40.4& 3.1& 5.8& 42.7& 5.2 &9.3& 43.8& 11.5& 18.2\\
     \midrule
    &PIFS&40.4& 10.4& 16.6& 40.1& 13.1& 19.8& 41.1& 18.3& 25.3\\
    &OINet&41.4&\textbf{11.7}&\textbf{18.2}&41.5&\textbf{14.4}&\textbf{21.4}&41.5&\textbf{19.7}&\textbf{26.7}\\
  \bottomrule
\end{tabular}
}
\caption{Results on COCO.}
  \label{result_ifss_coco}
\end{table*}
\vspace{-0.1cm}
\section{Experiments}
\subsection{Dataset}
We use Pascal-VOC 2012 (VOC) \cite{pascal_voc} containing 20 classes, and COCO \cite{coco_2,coco_24} where, as in previous works \cite{canet,pifs}, we use the 80 thing classes. We consider 15 and 60 of the classes as base $\left(\mathcal{C}^0\right)$ and 5 and 20 as new (novel) $\left(\mathcal{C}^t \backslash \mathcal{C}^0\right)$, for VOC and COCO respectively.
They are for segmentation task \cite{shan2021class,shan2021decouple,shan2021densenet,shan2021uhrsnet,shan2022class,shan2022mbnet,shan2023boosting,shan2023data,shan2023incremental,wu2023continual,zhao2023explore,zhao2023flowtext,zhao2023generative,zhao2024controlcap}.
The protocols start with pretraining on base classes and multiple steps on novel classes, i.e., 5 steps of 1 novel class on VOC and 4 steps of 5 novel classes on COCO.
In line with PIFS, we divide VOC in 4 folds of 5 classes and COCO in 4 folds of 20 classes, running experiments 4 times by considering each fold in turn as the set of new classes.
For each setting, we consider 1, 2, or 5 images in an incremental learning step, and we average the results over multiple trials, each using a different set of images. The images are randomly sampled from the set of images containing at least one pixel of the new classes without imposing any restriction on the existence of the old class. 
During incrementally learning novel classes, we only rely on the few images provided.
In learning the base classes, consistent with previous incremental semantic segmentation works \cite{mib,plop}, labels except base classes are marked as background, which best fits the actual world situation.
We use cross-validation, i.e., 20\% of the data in the training set is regarded as the real validation set.
Finally, we report results on each dataset's entire validation set (for the test), considering all visible classes, which is consistent with previous works \cite{pifs,lwf}.

\subsection{Implementaion Details}
Deeplab-v3 \cite{deeplabv3} with ResNet-101 \cite{resnet} is used for all experiments, and in-place batch normalization \cite{inplace_BN} is employed to reduce memory.
ResNet-101 with ASPP is the feature extractor, and a $1 \times 1$ convolutional layer is the classifier (segmentation head).
As it is standard practice in FSS and IL \cite{mib,fss_8}, we initialize ResNet using an ImageNet pre-trained model.
When fine-tuning, we follow \cite{mib}, using SGD as optimizer with momentum $0.9$, weight decay $10^{-4}$ and a polynomial learning rate policy, i.e. $1 \mathrm{r}=\operatorname{lr}_{\text {init}}\left(1-\frac{\text {iter}}{\max \text {iter}}\right)^{0.9}$.
During training, we apply the same data augmentation of \cite{deeplabv3}, performing random scaling and horizontal flipping with a crop size of $512 \times 512$.
In line with PIFS, we use a different learning rate and training iterations depending on the dataset, the number of shots, and the learning steps.
Specifically, during learning base classes, networks are trained for 30 epochs on Pascal-VOC and 20 epochs on COCO with a learning rate of $10^{-2}$ and batch size of 24.
During incrementally learning step $t$, we set the batch size to $\min \left(10,\left|\mathcal{D}_t\right|\right)$.
In the incremental training of VOC, the network is trained for 200 iterations per step with a learning rate of $10^{-4}$.
In COCO, the network is trained for 100 iterations per step with a learning rate of $10^{-4}$.
These hyperparameters are shared by all methods.
We compute the results via single-scale full-resolution images without any post-processing.

\subsection{Results}
Following the protocol GFSS work \cite{gfss} and PIFS, we evaluate the method's performances via three metrics based on the mean Intersection over Union (mIoU): mIoU on base classes (mIoU-Base), mIoU on novel classes (mIoU-Novel), and the harmonic mean of the two (HM). We report the results after the last step as in \cite{mib,ilt}. 

The comparison methods include fine-tuning (FT), few-shot classification (FSC), few-shot segmentation (FSS), incremental learning (IL), and the PIFS method for iFSS.
FSC includes weight imprinting (WI) \cite{wi}, dynamic imprinting (DWI) \cite{dwi}, and RT \cite{rt}.
FSS includes adaptive masked proxies (AMP) \cite{amp} and semantic projection networks (SPN) \cite{spn}.
IL methods include learning without forgetting (LwF) \cite{lwf}, incremental learning techniques (ILT) \cite{ilt}, and modeling the background (MiB) \cite{mib}.

\textbf{Results on Pascal-VOC:}
For novel classes, 1-shot, 2-shot, and 5-shot experiments are performed, respectively.
The results are shown in Table \ref{result_ifss_voc}. It can be seen that the FSC methods, especially WI and DWI, show very good results. And the effects on the base and novel classes are both competitive, which shows that the imprint method is very suitable for iFSS tasks.
However, for the methods of FSS and IL, the performances on both base classes and novel classes are very poor.
These manifestations are explainable.
FSS methods do not have any measure for keeping base class memory.
While IL methods need large samples for novel classes to learn them, the limited setting of few-shot makes the its role unable to play.
Compared with PIFS (the baseline of our method), our method has significant improvements both in the memory of base classes and the learning of novel classes, demonstrating our method's effectiveness.

\textbf{Results on COCO:}
The results on COCO are consistent with VOC, i.e., the methods of FSC perform well, while the methods of FSS and IL perform poorly.
The COCO data set is more challenging than VOC, so the performance of each method has a certain degree of decline.
However, our OINet achieves the best results no matter compared with the FSC methods or PIFS.



\subsection{Ablation Study}
\vspace{-0.1cm}
\begin{table}[H]
\centering
  \scalebox{0.86}{
  \begin{tabular}{ccc|cc|c|c|c}
Multi&$\mathcal{L}_{DSP}$&$\mathcal{L}_{CMP}$&DWI&$\mathcal{L}_{IHT}$&B&N&HM\\
    \toprule
    &&&&&64.1&16.9&26.7\\
    $\checkmark$&&&&&64.4&16.5&26.3\\
    $\checkmark$&$\checkmark$&&&&64.8&17.0&27.0\\
    $\checkmark$&$\checkmark$&$\checkmark$&&&65.3&17.0&26.9\\
     \midrule
     $\checkmark$&&&$\checkmark$&&64.9&17.3&27.3\\  
     $\checkmark$&&&$\checkmark$&$\checkmark$&65.2&17.5&27.6\\
      \midrule
$\checkmark$&$\checkmark$&$\checkmark$&$\checkmark$&&65.6&17.7&27.9\\  
$\checkmark$&$\checkmark$&$\checkmark$&$\checkmark$&$\checkmark$&65.8&18.0&28.3\\
    \bottomrule
\end{tabular}
}
\caption{Ablation of the different components.}
 \label{table_diff_losses}
\end{table}
\vspace{-0.1cm}

In order to verify the effectiveness of the proposed modules or optimization objectives, we conduct ablation experiments, and the results are shown in Table \ref{table_diff_losses}.
'Multi' represents the Multi sub-class prototypes of the background.
The module is divided into two parts: embedding space organization (including multi, $\mathcal{L}_{DSP}$ and $\mathcal{L}_{CMP}$ in the table) and inheritance mechanism (including DWI and $\mathcal{L}_{IHT}$).
For embedding space organization, it is helpful for the memory of the old class, but it has no obvious effect on learning novel classes.
When only using Multi, but no $\mathcal{L}_{DSP}$ and $\mathcal{L}_{CMP}$, the performance of the novel classes will even drop. This phenomenon may be due to the features of the background not being well compacted and dispersed, which causes a certain confusion with the features of the novel classes, resulting in some novel classes being mistakenly classified as the background.
The inheritance mechanism is to make models learn new classes with few samples. After adding DWI, even without $\mathcal{L}_{DSP}$ and $\mathcal{L}_{CMP}$, the performance is improved compared to the baseline.
After adding $\mathcal{L}_{CMP}$, $\mathcal{L}_{DSP}$ and $\mathcal{L}_{IHT}$, the effects are all improved and more obvious.
It is because after adding the compaction and dispersion operations, the influence of the inheritance on other irrelevant classes or prototypes can be reduced.
In a nutshell, embedding space organization can be regarded as the basic operation, and only on the basis the inheritance mechanism can its play the role to the maximum.
\vspace{-0.1cm}

\subsection{Sensitive Analysis}

\begin{table}[H]
\centering
  \scalebox{0.9}{
  \begin{tabular}{c|ccccc}
    \toprule
    \multicolumn{1}{c|}{K}&5& 10 & 15 & 20 & 25\\
    \midrule
    N=1&28.4&28.5&28.3&28.3&28.0\\
    \midrule
    N=2&27.7&27.9&28.3&28.4&28.1\\
    \midrule
    N=3&27.7&27.9&28.0&28.2&28.1\\
    \bottomrule
\end{tabular}
}
\caption{Ablative studies of sub-class prototypes $K$.}
 \label{table_proto_num}
\end{table}
\vspace{-0.2cm}
\textbf{Results with different sub-class prototypes:}
The important contribution of our work is to create embedding space for the learning of novel classes through the organization of sub-class prototypes in the background. Therefore, analyzing the impact of the different prototype numbers (i.e., K) is necessary.
Results are shown in Table \ref{table_proto_num}. N represents the number of executed times of the KM algorithm, i.e., assign N sub-class prototypes to each novel class.
It can be seen that the number of K has a certain influence on the final result. When the number of prototypes is small, and the prototypes generated by base training are occupied, the newly generated prototypes have limited effect due to less training data, which degrades performance.
When there are too many prototypes, the features belonging to one class may be collected into two clusters, resulting in a decline in results.
But whether it is more or less, the result obtained is much improved compared to not using sub-class prototypes, i.e., even the worst result is significantly improved compared to the baseline, demonstrating our work's significance.

\begin{table}[H]
\centering
  \scalebox{0.90}{
  \begin{tabular}{c|cccccc}
    \toprule
   & 0.001 & 0.005 & 0.01 & 0.02 & 0.03 & 0.05\\
    \midrule
     $\lambda_1$&28.1&28.3&28.3&28.2&28.0&27.8\\
    \midrule
     $\lambda_2$&27.7&27.9&28.3&28.5&28.5&28.5\\
    \midrule
     $\lambda_3$&28.2&28.3&28.3&28.0&27.6&27.4\\
    \bottomrule
\end{tabular}
}
\caption{Analysis of weights, $\lambda_1$, $\lambda_2$, and $\lambda_3$.}
 \label{table_diff_weight}
\end{table}
\vspace{-0.2cm}

\textbf{Analysis of $\lambda_1$, $\lambda_2$, and $\lambda_3$:}
Table \ref{table_diff_weight} shows the impact of the three hyper-parameters of the losses on the final result.
When one changes, the other two are fixed as default.
It can be seen that the model is robust to the three hyper-parameters, especially for $\lambda_1$ and $\lambda_2$. The effect will drop when $\lambda_3$ is too large because the newly generated classification heads cannot obtain enough guidance from the training data of the novel classes when $\lambda_3$ is too large. When $\lambda_3$ $\in$ [0.005, 0.01], the best result can be obtained.
\vspace{-0.2cm}

\begin{table}[H]
\centering
  \scalebox{0.90}{
  \begin{tabular}{c|cccccc}
    \toprule
    $\mu$& 0.9 & 0.99 & 0.999&0.9999\\
    HM&26.6&27.7&28.3&28.0\\
    \bottomrule
\end{tabular}
}
\caption{Analysis of momentum coefficient $\mu$.}
 \label{table_mu}
\end{table}
\vspace{-0.2cm}
\textbf{Analysis of update coefficient $\mu$:}
We use the prototype's gradual update strategy, and the effect of the momentum coefficient ($\mu$ in Equation (\ref{eq_mu})) on the final result should be verified. Results are shown in Table \ref{table_mu}.
The model performs well with relatively large coefficients (i.e. $\mu$ $\in$ [0.999,0.9999]), and performance drops when $\mu$ is $0.9$ or $0.99$.
When the coefficient is larger, the obtained information is more global and, therefore, more representative, thus achieving better results.

\section{Conclusion}
In this work, according to the characteristics of iFSS, we set up multiple sub-class prototypes for the background embedding space and perform intra-class (prototype) feature compaction and inter-class feature dispersion to open up space for novel class learning. When learning novel classes, we select the most matched prototypes from the multiple sub-class prototypes of background based on IoU. And their embedding spaces are used as the space of novel classes, which will not break the feature distribution of old classes but can facilitate the learning of novel classes. 
Our idea of considering the learning of novel classes when base training is worth learning.
In future work, how to better organize the background features, not just simple clustering, is a direction worthy of further exploration, which can take inspiration from self-supervised works.

\bibliographystyle{named}
\bibliography{ijcai23}

\end{document}